\documentclass[runningheads,a4paper,orivec]{llncs}

\usepackage[dvipsnames]{xcolor}
\usepackage{xspace}
\usepackage{tabularx}
\usepackage{multicol}
\usepackage{paralist}
\usepackage{wrapfig}
\usepackage[utf8]{inputenc}
\usepackage[T1]{fontenc}
\usepackage{color}
\usepackage[inline]{enumitem}
\usepackage{graphicx}
\usepackage{multirow}
\usepackage{bussproofs}
\usepackage{comment}
\usepackage{mathpartir}
\usepackage{textcomp}
\usepackage[super]{nth}

\usepackage{ebproof}

\usepackage{amsmath}
\usepackage{prettyref}%
\newrefformat{sec}{Section~\ref{#1}}%
\newrefformat{fig}{Figure~\ref{#1}}%
\newrefformat{tab}{Table~\ref{#1}}%

\usepackage{hhline}%

\usepackage{amsthm}
\usepackage{amssymb}

\DeclareMathOperator{\subformula}{sub} %
\DeclareMathOperator{\emb}{enc} %

\DeclareMathOperator*{\intersect}{intersect}
\DeclareMathOperator*{\disjoint}{disjoint}

\usepackage{tikz}%

\usetikzlibrary{shapes.misc, positioning}%
\usetikzlibrary{decorations.pathreplacing}%
\usetikzlibrary{fadings}%
\usetikzlibrary{positioning,shapes.geometric}%
\usetikzlibrary{matrix,arrows}%
\usetikzlibrary{calc}%

\usepackage{color, colortbl} \definecolor{Gray}{gray}{0.93}

\definecolor{darkblue}{HTML}{00007F}
\usepackage[bookmarks=true,colorlinks=true,allcolors=darkblue]{hyperref}

\begin{document}

\title{A Study of Continuous Vector Representations for Theorem Proving}

\author{Stanisław Purgał\inst{1} \and Julian Parsert\inst{2} \and Cezary Kaliszyk\inst{1,3}}
\authorrunning{Purgał, Parsert \and Kaliszyk}

\institute{
University of Innsbruck, Innsbruck, Austria\\
\email{\{stanislaw.purgal,cezary.kaliszyk\}@uibk.ac.at}
\and
University of Oxford, Oxford, UK\\
\email{julian.parsert@gmail.com} 
\and
University of Warsaw, Poland\\
}

\maketitle

\begin{abstract}
  Applying machine learning to mathematical terms and formulas requires
  a suitable representation of formulas that is adequate for AI methods.
  In this paper, we develop an encoding that allows for logical properties
  to be preserved and is additionally reversible. This means that the
  tree shape of a formula including all symbols can be reconstructed
  from the dense vector representation. We do that by training two decoders: one that extracts the top symbol of the tree and one that extracts embedding vectors of subtrees.
  The syntactic and semantic logical properties that we aim to preserve
  include both structural formula properties, applicability of natural deduction
  steps, and even more complex operations like unifiability.
  We propose datasets that can be used to train
  these syntactic and semantic properties. We evaluate the viability of the developed encoding across the proposed
  datasets as well as for the practical theorem proving problem of premise selection
  in the Mizar corpus.
\end{abstract}

\section{Introduction}\label{sec:intro}

The last two decades saw an emergence of computer systems applied to logic
and reasoning. Two kinds of such computer systems are
interactive proof assistant systems~\cite{HarrisonUW14} and automated
theorem proving systems~\cite{RobinsonV01}. Both
have for a long time employed human-developed heuristics and AI methods,
and more recently also machine learning components.

Proof assistants are mostly
used to transform correct human proofs written in standard mathematics
to formal computer understandable proofs. This allows for a verification of proofs
with the highest level of scrutiny, as well as an automatic extraction of additional information from the proofs. Interactive theorem provers
(ITPs) were initially not intended to be used in standard mathematics,
however, subsequent algorithmic developments and modern-day computers
allow for a formal approach to major mathematical proofs~\cite{hales2008formal}.
Such developments include the proof of Kepler's conjecture~\cite{hales2017-kepler}
and the four colour theorem~\cite{gonthier2008formal}. ITPs are also used to
formally reason about computer systems, e.g. have been used to develop a formally
verified operating system kernel~\cite{KleinAEHCDEEKNSTW10} and a verified
C compiler~\cite{Compcert09}. The use of ITPs is still more involved and requires
much more effort than what is required for traditional mathematical proofs. Recently,
it has been shown that machine learning techniques combined with automated reasoning
allow for the development of proofs in ITPs that is more akin to what we are used to in traditional mathematics~\cite{h4qed}.

Automated reasoning has been a field of research since the sixties. Most Automated Theorem
Proving systems (ATPs) work in less powerful logics than ITPs. They are most
powerful in propositional logic (SAT solvers), but also are very strong in classical
first-order logic. This is mostly due to a good understanding of the underlying calculus
and its variants (e.g. the superposition calculus for equality~\cite{BachmairGLS92}),
powerful low-level programming techniques, and the integration of bespoke heuristics and
strategies, many of which took years of hand-crafting~\cite{0001CV19,Voronkov14}.

In the last decade, machine learning techniques became more commonly
used in tools for specifying logical foundations and for reasoning.
Today, the most powerful proof automation in major interactive theorem
proving systems filter the available knowledge~\cite{kuehlwein/premise/selection}
using machine learning components (Sledgehammer~\cite{jbdgckdkju-jar-mash16}, CoqHammer~\cite{DBLP:journals/jar/CzajkaK18}).
Similarly, machine-learned knowledge selection techniques have been included in
ATPs~\cite{ckssjujv-cade15et}. More recently, techniques that actually use machine learning
to guide every step of an automated theorem prover have been considered~\cite{malecop,Loos}
with quite spectacular success for some provers and domains: A leanCoP strategy found
completely by reinforcement learning is 40\% more powerful than the best human
developed strategy~\cite{ckjuhmmo-nips18}, and a machine-learned E prover strategy can again prove more
than 60\% more problems than the best heuristically found one~\cite{ChvalovskyJ0U19enigma}. All these new results
rely on sophisticated characterizations and encoding of mathematics that are also suitable for learning methods.

The way humans think and reason about mathematical formulas is very different from
the way computer programs do. 
Humans familiarize themselves with
the concepts being used, i.e. the context of a statement. This may include auxiliary
lemmas, alternative representations, or definitions. In some cases, observations
are easier to make depending on the representation used~\cite{gonthier2013machine}.
Experienced mathematicians may have seen or proven similar theorems, which can be
described as intuition. 
On the other hand, computer systems derive facts by manipulating syntax according to
inference rules. Even when coupled with machine learning that tries to
predict useful statements or useful proof steps the reasoning engine has very little
understanding of a statement as characterized by an encoding. We believe this to be
one of the main reasons why humans are capable of deriving more involved theorems than
modern ATPs, with very few exceptions~\cite{KinyonVV13}. 

In this paper, we develop a computer representation of mathematical objects (i.e. formulas, theorem statements, proof states),
that aims to be more similar to the human understanding of formulas than the existing representations. Of course, human understanding
cannot be directly measured or compared to a computer program, so we focus on an approximation of human understanding as
discussed in the previous paragraph. In particular, we mean that
we want to perform both symbolic operations and ``intuitive steps'' on the representation. By symbolic operations, we mean basic logical inference steps, such as modus ponens,
and more complex logical operations, such as unification. When it comes to the more intuitive steps, we would like the
representation to allow direct application of machine learning to proof guidance or even conjecturing.
A number of encodings of mathematical objects as vectors have been implicitly created as part of deep learning approaches
applied to particular problems in theorem proving~\cite{IrvingSAECU16,wang2017premise,olk2019property}. However, none of
them have the required properties, in particular, the recreation of the original statement from the vector is mostly
impossible.

It may be important to already note, that it is impossible to perfectly preserve all the properties
of mathematical formulas in finite-length vectors of floating-point values. Indeed, there are
finitely many such vectors and there are infinitely many formulas. It is nonetheless very interesting
to develop encodings that will preserve as many properties of as many formulas as possible, as this
will be useful for many practical automated reasoning and symbolic computation problems.

\paragraph{Contribution}
We propose methods for supervised and unsupervised learning of an
encoding of mathematical objects. By encoding (or embedding) we mean
a mapping of formulas to a continuous vector space. We consider two
approaches: an explicit one, where the embedding is trained to preserve
a number of  properties useful in theorem proving and an implicit one,
where an autoencoder of mathematical expressions is trained. For this
several training datasets pertaining to individual logical properties
are proposed. We also test our embedding on a known automated theorem
proving problem, namely the problem of premise selection. We do so using the Mizar40 dataset \cite{Kaliszyk2015MizAR4F}. 
The detailed contributions are as follows:
\begin{itemize}
\item We propose various properties that an embedding of first-order logic
  can preserve: formula well-formedness, subformula property, natural
  deduction inferences, alpha-equivalence, unifiability, etc. and propose
  datasets for training and testing these properties.
\item We discuss several approaches to obtaining a continuous vector representation of logical formulas.
  In the first approach, representations are learned using logical properties (explicit approach), and the second approach is based on autoencoders (implicit approach).
\item We evaluate the two approaches for
  the trained properties themselves and for a practical theorem proving
  problem, namely premise selection on the Mizar40 dataset.
\end{itemize}
The paper extends our work presented at GCAI 2020~\cite{ParsertAK20}, which
discussed the explicit approach to training an embedding that preserves
properties. The new material in this version comprises an autoembedding of first-order logic
(this includes the training of properties related to decoding formulas), new neural network
models considered (WaveNet model and Transformer model), and a more thorough evaluation. In particular,
apart from the evaluation of the embeddings on our datasets, we also considered a practical
theorem proving problem, namely premise selection on a standard dataset.

\paragraph{Contents}

The rest of this paper is structured as follows. 
In \prettyref{sec:prelim} we introduce the logical and machine learning preliminaries.
In \prettyref{sec:related} we discuss related work.
In \prettyref{sec:approach} we present two methods to develop a reversible embedding: the explicit
approach where properties are trained together with the embedding and the implicit approach where
autoencoding is used instead.
In \prettyref{sec:datasets} we develop a logical properties dataset and present the Mizar40 dataset.
\prettyref{sec:eval} contains an experimental evaluation of our approach. Finally \prettyref{sec:concl}
concludes and gives an outlook on the future work.

\section{Preliminaries}\label{sec:prelim}

\subsection{Logical Preliminaries}

In this paper we will focus on first-order logic (FOL).
We only give a brief overview, for a more
detailed exposition see Huth and Ryan~\cite{DBLP:books/daglib/huth:ryan:lics}.

An abstract Backus-Naur Form (BNF) for FOL formulas is presented
below. The two main concepts are terms (\ref{eqs:bnf:terms}) and
formulas (\ref{eqs:bnf:formula}). A formula can either be an Atom
(which has terms as arguments), two formulas connected with a logical
connective, or a quantified variable or negation with a formula.
Logical connectives are the usual connectives negation, conjunction,
disjunction, implication and equivalence. In addition, formulas can
be universally or existentially quantified.
\begin{align}
  \text{term} &:= \text{var}\ |\ \text{const}\ |\ f(\text{term},\dots, \text{term})\label{eqs:bnf:terms}\\%
  \text{formula} &:= \text{Atom}(\text{term},\dots, \text{term}) \label{eqs:bnf:formula}\\%
              &\ \ \ \ |\ \lnot \text{formula}\ |\ \text{formula} \land \text{forumla} \nonumber \\%
              &\ \ \ \ |\ \text{formula} \lor \text{formula}\nonumber \\%
              &\ \ \ \ |\ \text{formula} \to \text{formula}\ |\ \text{formula} \leftrightarrow \text{formula} \nonumber\\%
              &\ \ \ \ |\ \exists\ \text{var}.\ \text{formula}\ |\ \forall\ \text{var}.\ \text{formula} \nonumber
\end{align}
For simplicity we omitted rules for bracketing. However, the
``standard'' bracketing rules apply. Hence, a formula is well-formed
if it can be produced by~(\ref{eqs:bnf:formula}) together with the
mentioned bracketing rules. The implementation is based on the syntax
of the FOL format used in the ``Thousands of Problems for Theorem
Provers'' (TPTP) library~\cite{Sut17}\footnote{The full BNF is available at: \texttt{\url{http://www.tptp.org/TPTP/SyntaxBNF.html}}
}. This library is very diverse as it contains data from various
domains including set theory, algebra, natural language processing and
biology all expressed in the same logical language. Furthermore, its
problems are used for the annual CASC competition for automated
theorem provers. Our data sets are extracted from and presented in
TPTP's format for first-order logic formulas and terms. An example for
a TPTP format formula is \texttt{![D]: ![F]: (disjoint(D,F) <=>
  \texttildelow intersect(D,F))} which corresponds to the formula
$\forall d.\ \forall f.\ \disjoint(d,f) \iff \neg \intersect(d,f)$. As
part of the data extraction, we developed a parser for TPTP formulas
where we took some liberties. For example, we allow for occurrences of
free variables, something the TPTP format would not allow.

To represent formulas we use labeled, rooted trees. So every node in our trees has some \textit{label} attached to it, and every tree has a special \textit{root} node. We refer to the label of the root as the \textit{top symbol}.

\subsection{Neural Networks}
Neural networks are a widely used machine learning tool for approximating complicated functions. In this work, we experiment with several neural architectures for processing sequences.

\begin{figure}[ht]
    \centering
    \includegraphics[height=5cm]{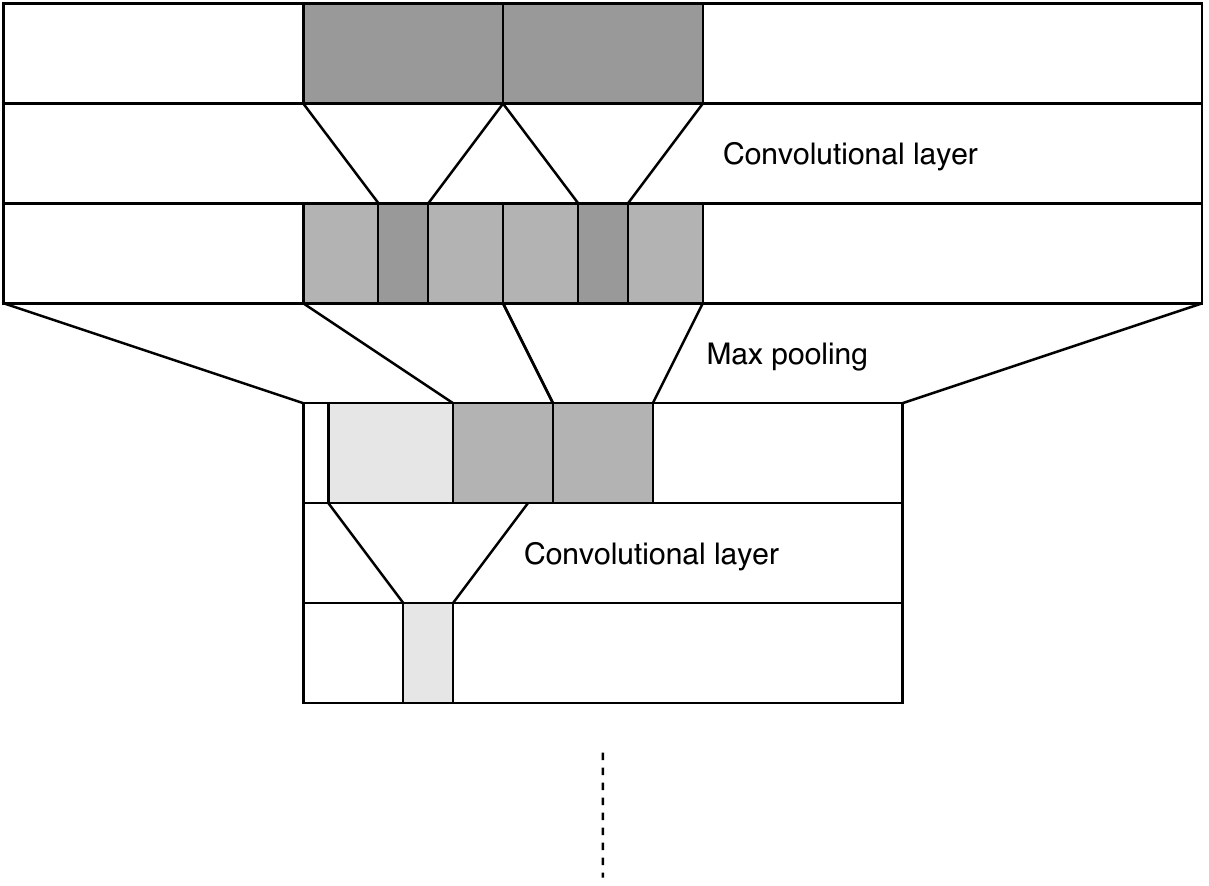}
    \caption{Convolutional network}
    \label{fig:conv}
\end{figure}

\paragraph{Convolutional Neural Networks}
Convolutional neural networks (\prettyref{fig:conv}) are widely used in computer vision \cite{10.1145/3065386} where they usually perform two-dimensional convolutions. However, in our case, the input of the network are string representations of formulas, which is a one-dimensional object.  Therefore,
we only need one-dimensional convolutions.

\begin{figure}[ht]
    \centering
    \includegraphics[height=5cm]{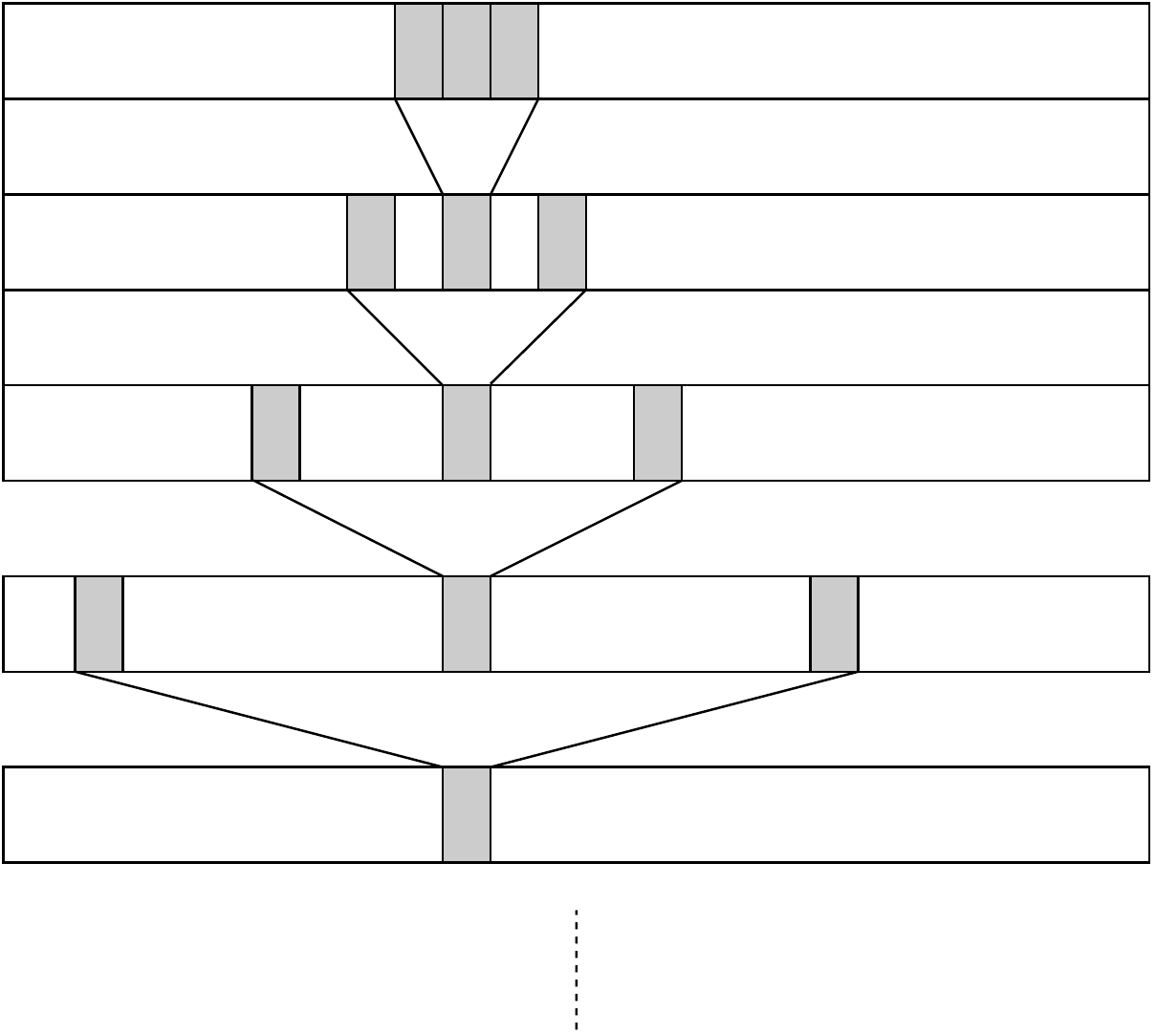}
    \caption{WaveNet network}
    \label{fig:wavenet}
\end{figure}

\begin{figure}[ht]
    \centering
    \includegraphics[height=5cm]{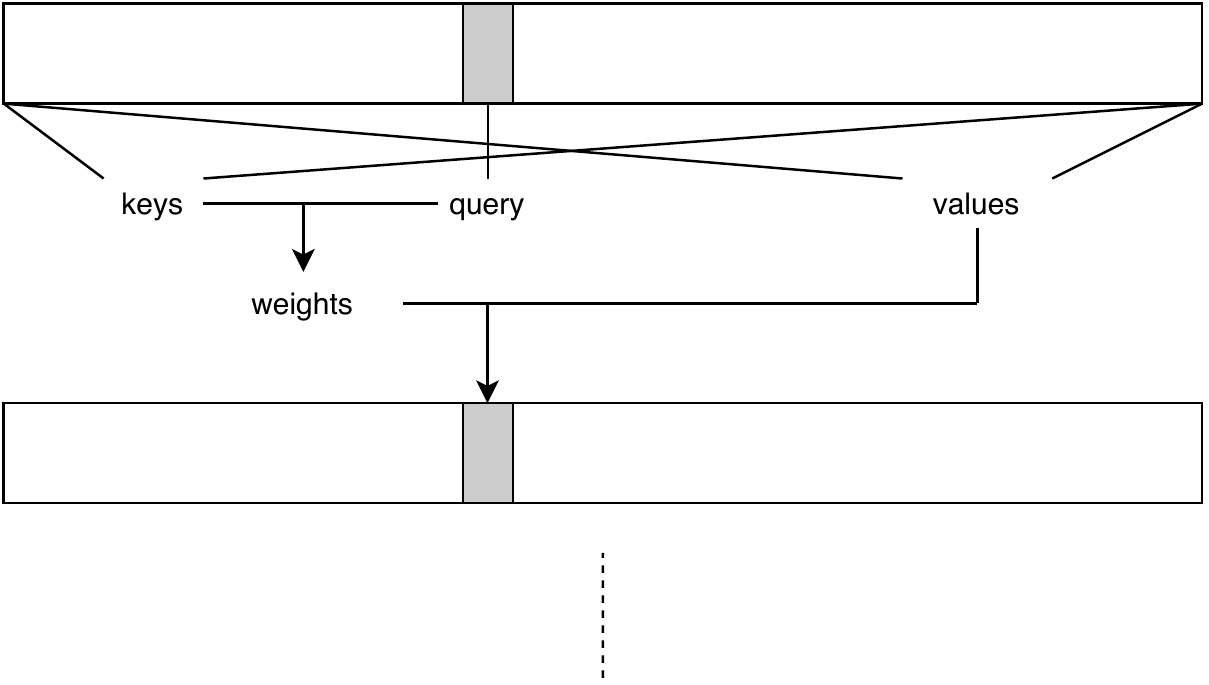}
    \caption{Transformer encoder network}
    \label{fig:transformer}
\end{figure}

In this kind of network, convolutional layers are usually used together with spatial pooling, which reduces the size of the object by aggregating several neighbouring cells (pixels or characters) into one. This is illustrated in \prettyref{fig:conv}.
\paragraph{Long-Short Term Memory}
Long-Short Term Memory networks~\cite{Hochreiter1997LongSM} are recurrent neural networks -- networks that process a sequence by updating a hidden state with every input token.
In an LSTM~\cite{Hochreiter1997LongSM} network, the next hidden state is computed using a forget gate, which in effect makes it easier for the network to preserve information in the hidden state.
LSTMs are able to learn order dependence, thanks to the ability to retain information long term, while at the same time passing short-term information between cells.
A \textit{bidirectional} network~\cite{bidir650093} processes sequences to directions and combines the final state with the final output.
\begin{figure}[ht]
    \centering
    \includegraphics[height=5cm]{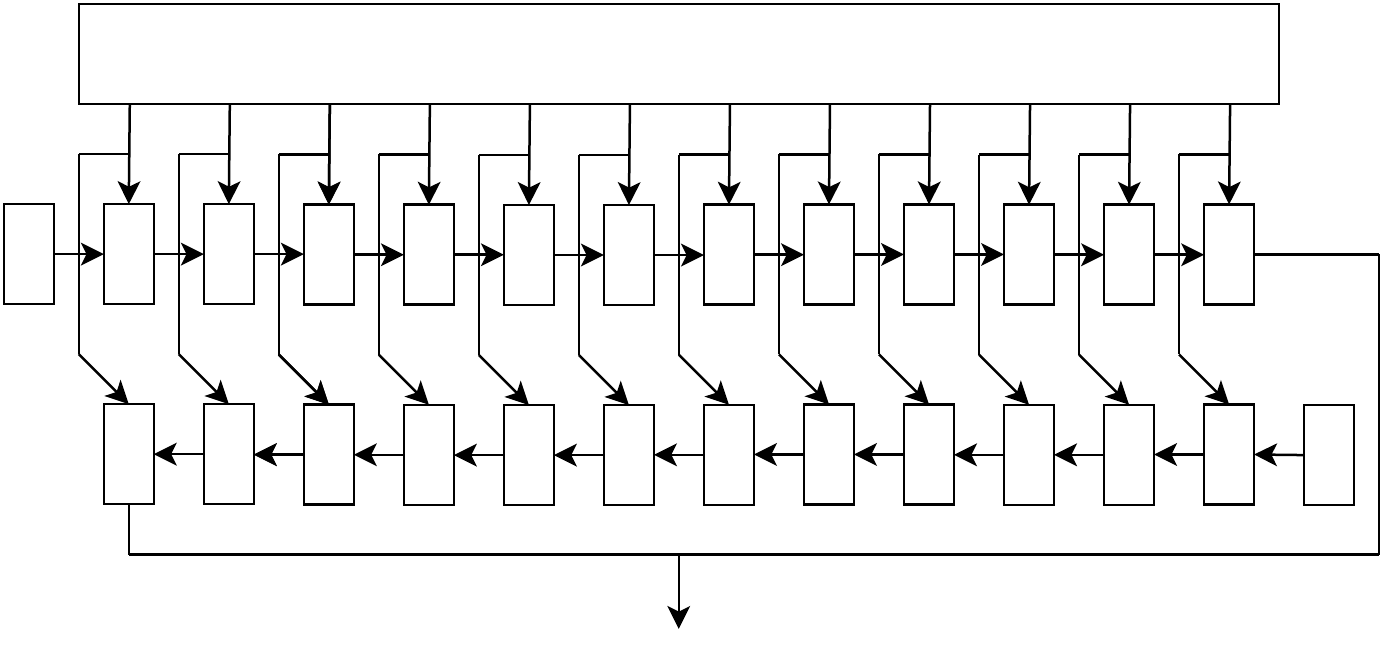}
    \caption{Bidirectional LSTM network}
    \label{fig:lstm}
\end{figure}
%
\paragraph{WaveNet}
WaveNet \cite{oord2016wavenet} is also a network based on convolutions. However, it uses an exponentially increasing dilation. That means that the convolution layer does not gather information from cells in the immediate neighbourhood, but from cells increasingly further away in the sequence. \prettyref{fig:wavenet} illustrates how the dilation increases the deeper in the network we are. This allows information to interact across large (exponentially large) distances in the sequence (i.e. formula). This kind of network performed well in audio-processing \cite{oord2016wavenet}, but also in proof search experiments \cite{Loos}. 
\paragraph{Transformer}
Transformer networks have been successfully applied to natural language processing~\cite{vaswani2017attention}. These networks consist of two parts, an encoder, and a decoder. As we are only interested in encoding we use the encoder architecture of a Transformer network~\cite{vaswani2017attention}. This architecture uses the attention mechanism to allow the exchange of information between every token in the sequence. An attention mechanism first computes \textit{attention weights} for each pair of interacting objects, then uses a weighted average of their embeddings to compute the next layer. In Transformer, the weights are computed as dot-product of ``key'' and ``query'' representations for every token. This mechanism is illustrated in \prettyref{fig:transformer}.

\paragraph{Autoencoders} \cite{doi:10.1002/aic.690370209} are neural networks trained to express identity function on some data. Their architecture usually contains some \textit{bottleneck}, which forces the network to learn patterns present in the data, to be able to reconstruct everything from smaller bottleneck information. This also means that all information about the input needs to be somehow represented within the bottleneck, which is the property we use in this work.

\section{Related work}\label{sec:related}

The earliest application of machine learning to theorem provers started in
the late eighties. Here we discuss only the deep-learning-based approaches
that appeared in recent years. As neural networks started being used for symbolic reasoning,
specific embeddings have been created for particular tasks. Alemi et al.~\cite{IrvingSAECU16} have first
shown that a neural embedding combined with CNNs and LSTMs can perform better than
manually defined features for premise selection. In a setup that also included
the WaveNet model, it was shown that formulas that arise in the automated theorem prover E
as part of its given clause algorithm can be classified effectively, leading to proofs found more efficiently~\cite{Loos}.

Today, most neural networks used for mathematical formulas are variants of Graph Neural Network \cite{hamilton2017inductive} -- a kind of neural network that repeatedly passes messages between neighbouring nodes of a graph. This kind of network is applied to the problem of premise selection by Wang et al. \cite{wang2017premise}. Later work of Paliwal et al. \cite{Paliwal2020GraphRF} experimented with several ways of representing a formula as a graph and also
consider higher-order properties.

A most extreme approach to graph neural networks for formulas was considered in \cite{olk2019property}, where a single hypergraph is constructed of the entire dataset containing all theorems and premises. In this approach, the symbol names are forgotten, instead, all references to symbols are connected within the graph. This allows constructing the graph and formulate message passing in a way that makes the output of network invariant under reordering and renaming, as well as symmetric under negation. A different improvement was recently proposed by Rawson and Reger~\cite{RawsonR19}, where the order of function and predicate arguments is uniquely determined by asymmetric links in the graph embedding.

The work of \cite{Crouse2019ImprovingGN} also uses graph neural networks with message passing, but after applying this kind of operation they aggregate all information using a Tree LSTM network \cite{Tai_2015}. This allows for representing variables in formulae with single nodes connected to all their occurrences, while also utilizing the tree structure of a formula. 
A direct comparison with works of this kind is not possible, since in our approach we explicitly require the possibility of decoding the vector back into formulas, and the other approaches do not have this capability.

Early approaches trying to apply machine learning to mathematical formulas have focused on manually defining feature spaces. In certain domains manually designed feature spaces prevail until today. Recently Nagashima~\cite{Nagashima19} proposed a domain-specific language for defining features of proof goals (higher-order formulas) in the interactive theorem prover Isabelle/HOL and defined more than 60 computationally heavy but useful features manually. The ML4PG framework~\cite{ml4pgii} defines dozens of easy to extract features for the interactive theorem prover Coq. A comparison of the different approaches to manually defining features in first-order logic together with features that rely on important logical properties (such as anti-unification) was done by the last author~\cite{ckjujv-ijcai15}.  Continuous representations have also been proposed for simpler domains, e.g. for propositional logic and for polynomials by Allamanis et al.~\cite{AllamanisCKS17}.

We are not aware of any work attempting to auto-encode logical formulas. Some efforts were however done to reconstruct a formula tree. Gauthier~\cite{gauthier-lpar2020} trained a tree network to construct a new tree, by choosing one symbol at a time, in a manner similar to sequence-to-sequence models. Here, the network was given the input tree, and the partially constructed output tree and tasked with predicting the next output symbol in a way similar to Tree2Tree models~\cite{tree2tree}. Neural networks have also been used for translation from informal to formal mathematics, where the output of the neural network is a logical formula. Supervised and unsupervised
translation with Seq2Seq models and transformer models was considered by Wang et al.~\cite{qwcbckju-cpp20,qwckju-cicm18}, however there the language considered as input was natural language. As such it cannot be directly compared to our current work that autoencodes formulas.
Autoencoder-based approaches have also been considered for programming language code, in particular, the closest to the current work was proposed by Dumančić et al.~\cite{DumancicGMB19} where Prolog code is autoencoded and operations on the resulting embedding are compared to other constraint solving approaches.


In natural language processing, pre-training on unsupervised data has achieved great results in many tasks \cite{mikolov2013distributed,Devlin2019BERTPO}. Multiple groups are working on transferring this general idea to informal mathematical texts, mostly by extending it to mathematical formulas in the ArXiv~\cite{YoussefM18}. This is, however, done by treating the mathematical formulas as plain text and without taking into account any specificity of logic.

\section{Approach}\label{sec:approach}
As previously mentioned, our main objective is an encoding of logical formulas. In particular, we are interested in networks
that take the string representation of a formula as an input and return a continuous vector representation thereof. This representation
should preserve properties and information that is important for problems in theorem proving. We considered two approaches, an implicit and an explicit approach. In the explicit approach, we defined a set number of logical properties (c.f. \prettyref{sec:eval-logical-properties}) and related classification problems and trained an encoding network with the loss of these classifiers. The implicit approach is based on autoencoders where we train a network that given a formula encodes it and then decodes it back to the same formula. In theory, this means that the encoding (i.e. continuous vector representation) preserves enough information to reconstruct the original formula. In particular, this means that
the tree structure of a formula is learned from its string representation. We will now explain the two approaches in detail starting with the explicit one.

\subsection{Explicit Approach}\label{sec:framework}
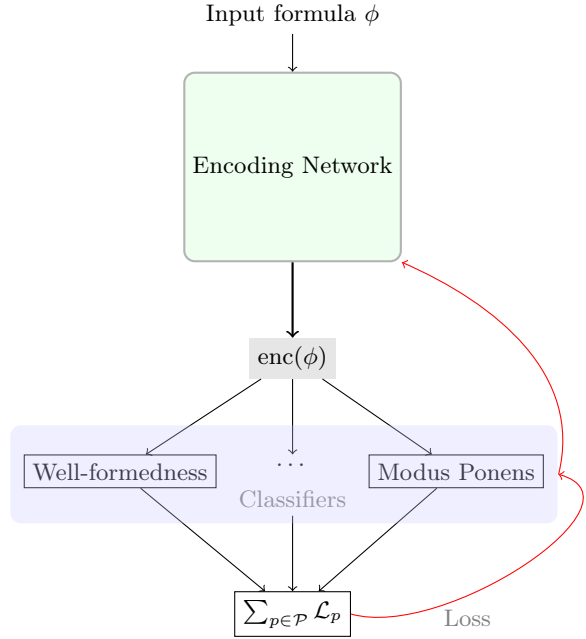
\begin{figure}[htb]
  \centering
  \begin{tikzpicture}
    \node[] (1) {Input formula $\phi$};%
    \node[draw] (2) [below=0.5 cm of 1] [rounded corners,
    opacity=0.30, fill=green!20, text opacity = 1, thick, minimum
    width=0.6cm, minimum height = 2.5cm] {Encoding Network};%
    \node[rectangle,fill=gray!20] (3) [below =of 2.south]
    {$\emb(\phi)$};%
    \node[rectangle,draw] (5) [ below right = of 3.south] {Modus
      Ponens};%
    \node[rectangle,draw] (6) [below left = of 3.south]
    {Well-formedness};%

    \node (7) [below = of 3.south] {\dots};%
    \node[opacity=0.5] (8) [below = 0.1cm of 7] {Classifiers};%
    \node[rectangle,draw] (9) [below = of 8]
    {$\sum_{p \in \mathcal{P}}{\mathcal{L}_p}$};%

    \draw [->] (5) -- (9);%
    \draw [->] (6) -- (9);%

    \draw [->] (1) -- (2);%
    \draw [->,thick] (2) -- (3);%
    \draw [->] (3) -- (5);%
    \draw [->] (3) -- (6);%
    \draw [->] (3) -- (7);%
    \draw [->] (8) -- (9);%
    \node[] (10) [rounded corners, opacity=0.3, fill=blue!20, below
    left = 0.6cm and -3.5cm of 3.south,thick, minimum width=7.2cm,
    minimum height = 1.3cm] {};%

    \draw [->,red] (9.east) to [bend right, out=-50, in=-50]
    (10.east);%
    \draw [->,red] (10.east) to [bend right, out=-50] (2.south east);%
    \node[opacity=0.5] (8) at (2.3,-8) {Loss};%
  \end{tikzpicture}
  \caption{The property training framework. The
    bottom area contains the classifiers that get one or more
    continuous representations of formulas $\emb(\phi)$ as input. If
    the classifier takes two formulas as input (i.e.
    alpha-equivalence), we gather $\emb(\phi_1)$ and
    $\emb(\phi_2)$ separately and forward the pair
    $(\emb(\phi_1),\emb(\phi_2))$ to the classifier. The encoding networks
    are described subsequently (cf. \prettyref{fig:models}).}
  \label{fig:learningFramework}
\end{figure}

The general setup for this approach is depicted in \prettyref{fig:learningFramework}.
The green box in \prettyref{fig:learningFramework} represents an encoding network for which we
consider different models which we discuss later in this section.
This network produces an encoding $\emb(\phi)$ of a formula $\phi$. This
continuous vector representation is then fed into classifiers that
recognise logical properties (c.f. \prettyref{sec:logical-properties}). The
total loss $\mathcal{L}$ is calculated by taking the sum of the losses
$\mathcal{L}_P$ of each classifier of the properties
$p \in \mathcal{P}$ discussed before. $\mathcal{L}$ is then propagated
back into the classifiers and the encoding network. This setup is
end-to-end trainable and ensures, that the resulting embedding
preserves the properties discussed in \prettyref{sec:logical-properties}.
We train the network on this setup and evaluate the whole training
setup (encoding network and classifiers) on unseen data in
\prettyref{sec:eval}. However, it is important to note that we are only interested in the encoding network.
Hence, we can extract the encoding network (c.f.
\prettyref{fig:learningFramework}) and discard the classifiers after
training and evaluation. A drawback of this explicit method is that
we are working under the assumption that the logical properties that we select
are \emph{sufficient} for the tasks that the encodings are intended for in the end.
That is, the encodings may only preserve properties that are helpful in classifying the trained
properties but not further properties that the network is not trained with.
Hence, if the encodings are used for tasks that are not related to the logical properties
that the classifiers are trained with, the encodings may be of no use.

\begin{figure}
  \centering
  \begin{tikzpicture}[%
    ->, shorten >=2pt, >=stealth, node distance=0.3cm,
    noname/.style={%
      rectangle, minimum width=3cm, draw, font=\scriptsize, }
    ,scale=0.8]%
    \node[] (1) {Input $\phi$}; %
    \node[draw, noname ] (2) [below=of 1] {Embedding Layer};%
    \node[draw, noname ] (3) [below=of 2] {Convolution};%
    \node[draw, noname ] (4) [below=of 3] {Pooling};%
    \node[] (5)[below=of 4] {\dots};%
    \node[draw, noname ] (6) [below=of 5] {Convolution};%
    \node[draw, noname ] (7) [below=of 6] {Pooling};%
    \node[dashed,noname] (8) [below=of 7] {Fully Connected Layer(s)};%
    \node[dashed,noname] (proj) [below=of 8] {Projection Layer};%
    \node[dashed, noname] (a8)[right=of 8.east] {Fully Connected
      Layer(s)};%
    \node[dashed, noname] (projLSTM)[below=of a8.south] {Projection
      Layer};%

    \node[draw,noname, minimum height=3.2cm] (a4) [above=of a8.north]
    {Bidirectional LSTM layers};%
    \node[draw,noname] (a2) [above=of a4.north] {Embedding Layer};%
    \node[] (a1) [above = of a2.north] {Input $\phi$};%
    \node[] (9) [below = of proj] {$\emb(\phi)$};%

    \node[] (classifiersLSTM) [below =1.07cm of a8.south] {
      $\emb(\phi)$};%

    \path (1) edge node {} (2) (2)%
    edge node {} (3) (3) %
    edge node {} (4) (4) %
    edge node {} (5) (5) %
    edge node {} (6) (6) %
    edge node {} (7) (7) %
    edge node {} (8) (8) %
    edge node {} (proj) (proj) %
    edge node {} (9) (9); %
    \path (a1) edge node {} (a2) (a2) %
    edge node {} (a4) (a4) %
    edge node {} (a8) (a8) %
    edge node {} (projLSTM) (projLSTM) %
    edge node {} (classifiersLSTM) (classifiersLSTM);%
    \node[] (10) at (2.3,-4.3) [rounded corners, opacity=0.3,
    fill=green!20, thick, minimum width=7.8cm, minimum height = 6cm]
    {};%
  \end{tikzpicture}
  \caption{The encoding models we considered with the layers that the
    input passes through. The left diagram depicts CNN-based models,
    while the right one depicts LSTM-based models. The dashed
    boxes describe layers that are optional for these model types.}
  \label{fig:models}
\end{figure}
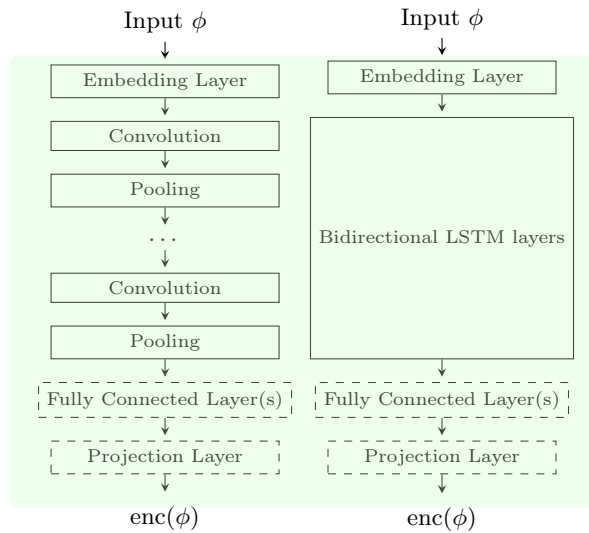

\paragraph{Classifiers}
The classifiers' purpose is to train the encoding network. 
This is implemented by jointly training the encoding
networks and classifiers.
There are two philosophies that can go into designing these classifiers. The
first is to make the classifiers as simple as possible, i.e. a single
fully connected layer. This means that in reality, the classifier can merely
select a subspace of the encoding. This forces the encoding networks to encode
properties in a ``high-level'' fashion. This is
advantageous if one wants to train simpler machine learning models with
the encodings. On the other hand, when using multiple layers in the
classifiers more complex relationships can be recognised by the classifiers
and the encoding networks can encode more complex features without
having to keep them ``high-level''. In this scenario, however, if the problems for the
classifiers are too easy it could happen that only the classifier
layers are trained and the encoding network layers remain
``untouched'' i.e. do not change the char-level encoding
significantly. We chose a middle ground by using two fully
connected layers, although we believe that o e could investigate
further solutions to this problem (e.g. adding weights to loss).
\paragraph{Encoding Models}\label{sec:explicit-approach-models}
We considered 20 different encoding models. However, they can be grouped
into ten CNN based models and ten LSTM based models. We varied
different settings of the models such as embedding dimension, output dimensions as well as
adding an additional fully connected layer.
The layouts of the two model types are roughly depicted in \prettyref{fig:models}.
The exact dimensions and sizes of the models
are discussed in \prettyref{sec:eval}.
\paragraph{CNN based models}
The models based on CNNs are depicted on the left in \prettyref{fig:models}.
The first layer is a
variable size embedding layer, the size of which can be changed. Once the formulas have been embedded, we pass
them through a set of convolution and (max) pooling layers. In our
current model, we have 9 convolution and pooling layers with increasing
filter sizes and ReLUs as activation functions. The output of the
final pooling layer comprises the encoding of the input formula. In
the second model, we append an additional set of fully connected layers
after the convolution and pooling layers. However, these do not reduce
the dimensionality of the vector representation. For that, we introduce
a third type of models, which we call embedding models. In embedding models, the
last layer is a projection layer which we tested with output
dimensions 32 and 64. Note that between the last pooling layer and the
projection layer one can optionally add fully connected layers like in
the previous model. In \prettyref{sec:eval} we evaluate these models.
\paragraph{LSTM based models}
The LSTM based models are depicted on the right side in \prettyref{fig:models}. Much like in previous models, the first layer
is an embedding layer. The output of which gets fed into
bidirectional LSTM layers. The output of these layers serves as the
encoding of our input formulas. As with the CNN based models, we also
considered models where an additional set of fully connected or
projection layers is added.
\subsection{Implicit Approach}
\begin{figure}[htb]
    \centering
    \includegraphics[height=6cm]{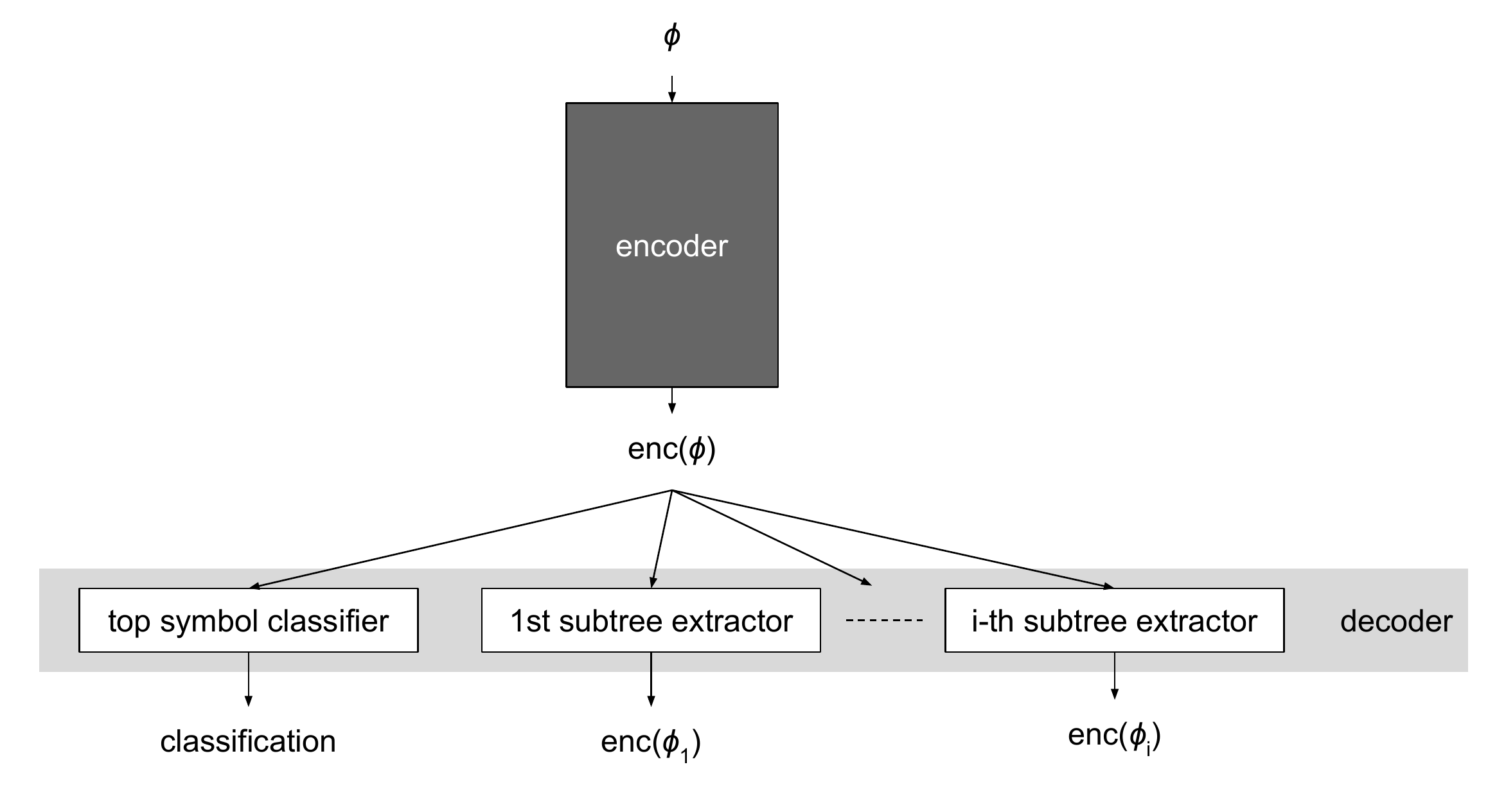}
    \caption{Tree autoencoder mechanism}
    \label{fig:tree_autoencoder}
\end{figure}

As previously mentioned the implicit approach does not work with specific logical properties. We use autoencoders to
encode formulas and subsequently retrieve the original formula from the encoding. As such the encoding has to contain
enough information about the original formula to reconstruct it from the encoding. Therefore, this method eliminates 
one of the major drawbacks of the previous approach where the encodings are dependent on the selected logical properties.
\prettyref{fig:tree_autoencoder} depicts a high-level overview of this setup.

We want to train the encoder to generate such continuous vector encodings that can be decoded. For this, we want the possibility to extract top symbol of a formula, as well as the encodings of all its subformulas. These two qualities would indeed enforce the encoding having the complete information about the entire tree-structure of a formula.

To achieve that, we train a top symbol classifier and subtree extractors together with the encoder.
The top symbol classifier is a single layer network that given the
encoding of a tree classifies it by its top symbol.
The subtree extractors are single linear transformations that output an encoding of the $i$-th  subtree.
Both encoders and decoders are trained together end-to-end using unlabelled data.
As with the explicit approach, we are not interested in decoder networks, and only use them to force the encoder to extract all information from the input.
The data (formulas) is provided in a string form but we require the ability to parse this data into trees.
\subsubsection{Difference training}

\begin{figure}[htb]
    \centering
    \includegraphics[height=6cm]{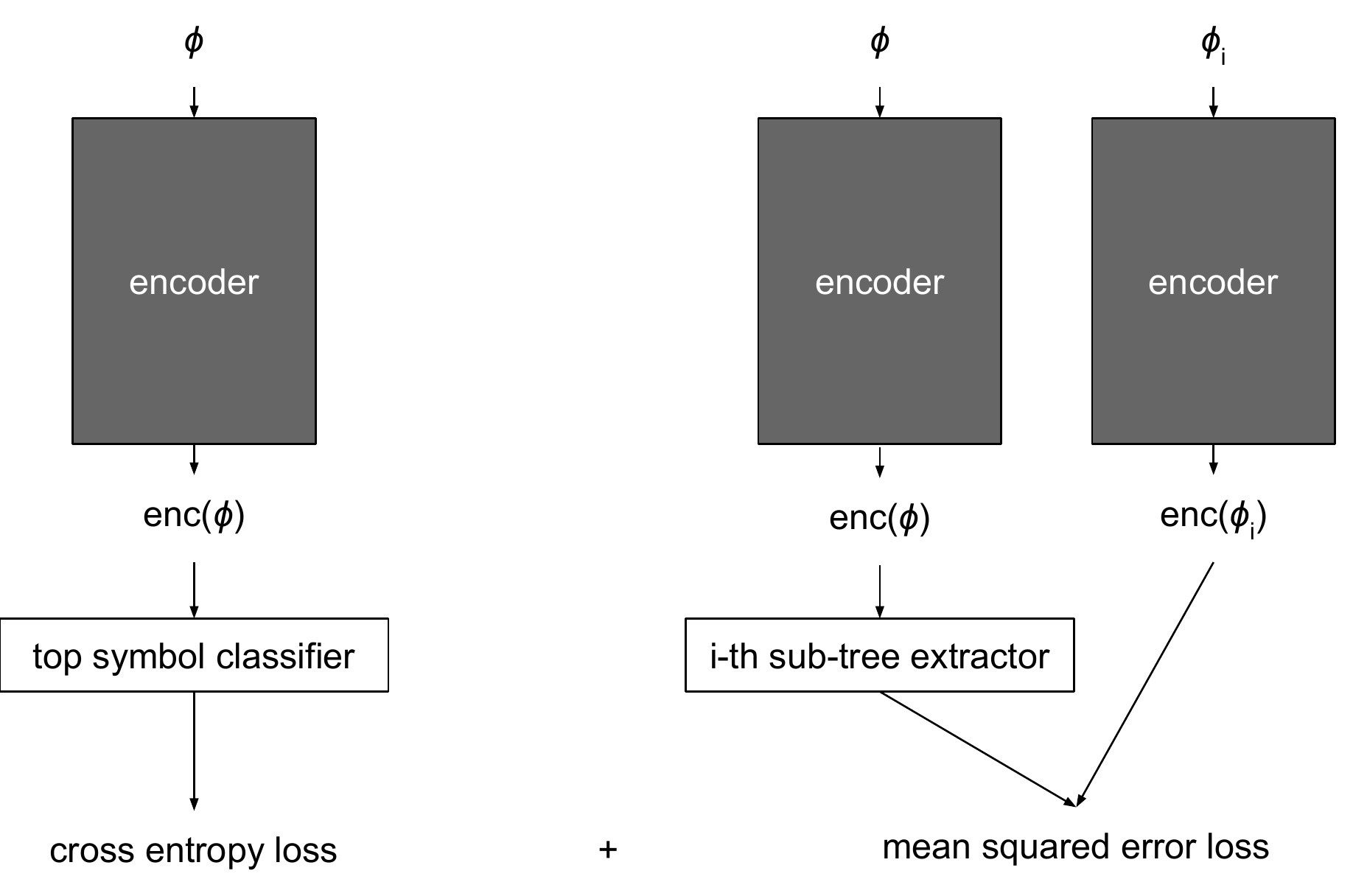}
    \caption{Difference training mechanism}
    \label{fig:diff_training}
\end{figure}

%
Our first approach is to train the top symbol classifier using cross-entropy loss and subtree extractor on mean square error loss using a dataset of all input trees and all their subtrees (\prettyref{fig:diff_training}). 

The first loss is forcing the embedding to contain information about the top symbol, and the second is about the subtrees. In the second loss, we force the result of extracting a subtree to be equal to the embedding of a subtree itself. Because of this, we need an encoding of the subtree by itself, and for this, we need the input string of a subtree. In formulae datasets, this is generally easy to achieve.

This method of training can be viewed as training on two datasets simultaneously. One dataset consists of formulas with their top symbol, and the other consists of formulas with their $i$-th subformula and the index $i$. The first dataset makes sure that the embedding of a formula contains information about its top symbol, and the second one makes sure that the embedding contains information about the embedding of all its subformulas. Together, those requirements force the embedding to contain information about the entire formula, in a form that is easily extracted with linear transformations.

Theoretically minimizing this loss enforces the ability to reconstruct the tree, however, given a practical limit on the size of the encoding, reconstruction fails above a certain tree depth. We do need to restrict the size of the encoding to one that will be useful for practical theorem proving tasks, like premise selection, etc. With such reasonable limits, we will later in the paper see that we can recover formulas of depth up to about 5, which is a very significant part of practical proof libraries.

\subsubsection{Recursive training}
\begin{figure}
    \centering
    \includegraphics[height=7cm]{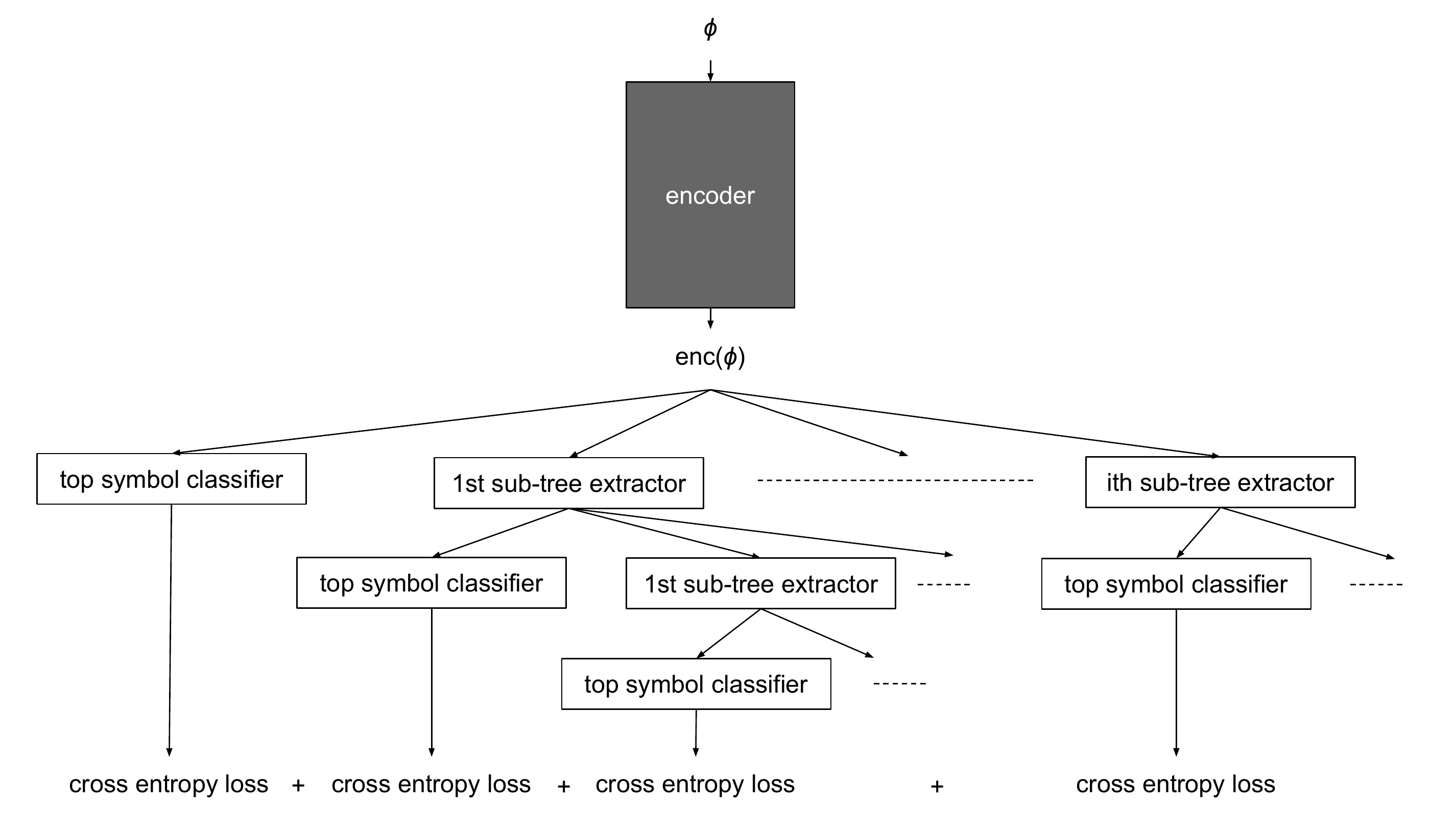}
    \caption{Recursive training mechanism}
    \label{fig:rec_training}
\end{figure}
In this method we only use the cross-entropy loss on top symbol classification. We compute encodings of subtree recursively (using subtree extractor transformations) and classify their top symbols as well (and so on recursively). All classification losses from a batch are summed together into one total loss that is used for back-propagation.

This is similar to tree recursive neural networks~\cite{Goller_1996}, like Tree LSTM \cite{Tai_2015} except pushing information in the other direction (from root to leaves) -- we reconstruct the tree from embedding and get a loss in every node.

In this approach, gradient descent can learn to recognize top symbols of subtrees even deep down the input tree. It is however much harder to properly parallelize this computation, making it much less efficient.

\subsubsection{Encoders}\label{sec:model}
As described above the encoding network is independent of the training setup. That is for both, difference and recursive training
different encoding models can be used. This is similar to the explicit approach where we also consider different encoding networks.
Here, in addition to the already considered \emph{CNN}s and \emph{LSTM}s, we will also consider \emph{WaveNet} and \emph{Transformer} models 
(introduced in \prettyref{sec:prelim}).

All these models receive as input a text string representation of a formula (a character level learned embedding). As output, they all provide a high-dimensional vector representation of a formula.
\section{Datasets}\label{sec:datasets}

We will consider two datasets for our training and for the experiments.
The first one is a dataset used to train logical properties, that we 
believe a formula embeddings should preserve. The dataset is 
extracted from TPTP. TPTP is a database of problems stated in first-order logic. It contains first-order problems from graph theory, category theory, and set theory among other fields.
These datasets differ in the problems themselves as well as vocabulary that is used to state said problems. For instance, in the set theory problem set one would find predicates such as \verb|member|, \verb|subset|, and \verb|singleton| whereas in the category theory dataset has predicates such as \verb|v1_funct_2|, and \verb|k12_nattra_1|.
The second dataset is the Mizar40 dataset \cite{Kaliszyk2015MizAR4F}, a known premise selection dataset. The neural network training part of the dataset consists of pairs of theorems and premises together with their statements, as well as the information
if the premise was useful in the proof or not. Half are positive examples and half are negatives. 

\subsection{Logical properties dataset}\label{sec:logical-properties}
We introduce some properties of formulas that we will consider 
in subsequent sections and describe how
the data was extracted.

\paragraph{Well-formedness:} As mentioned above it is important that
the encoding networks preserve the information of a formula being
well-formed. The data set was created by taking TPTP formulas as
positive examples and permutations of the formulas as negative
examples. We generate permutations by randomly iteratively swapping
two characters and checking if the formula is well-formed, if it is
not, we use it as a negative example. This ensures that the
difference between well-formed formulas and non well-formed formulas
is not too big.

\paragraph{subformula:} Intuitively, the subformula relation maps
formulas to a set of formulas that comprise the original formula.
Formally, the subformula relation is defined as follows:
\begin{gather*}\label{math:formula:bnf}
  \subformula(\phi) =\
  \begin{cases}
    \{ \phi \} &\text{if}\ \phi \text{ is}\ \text{Atom}\\ %
    \subformula(\psi) \cup \{\phi \} &\text{if}\ \phi\ \text{is}\ \neg
    \psi\\%
    \subformula(\psi_1) \cup \subformula(\psi_2) \cup
    \{\phi\}&\text{if}\ \phi\ \text{is}\ \psi_1 \land \psi_2\\%
    \subformula(\psi_1) \cup \subformula(\psi_2) \cup
    \{\phi\}&\text{if}\ \phi\ \text{is}\ \psi_1 \lor \psi_2\\%
    \subformula(\psi_1) \cup \subformula(\psi_2) \cup
    \{\phi\}&\text{if}\ \phi\ \text{is}\ \psi_1 \to \psi_2\\%
    \subformula(\psi_1) \cup \subformula(\psi_2) \cup
    \{\phi\}&\text{if}\ \phi\ \text{is}\ \psi_1 \leftrightarrow
    \psi_2\\%
    \subformula(\psi) \cup \{\phi\} &\text{if}\ \phi\ \text{is}\
    \forall x.\ \psi\\%
    \subformula(\psi) \cup \{\phi\} &\text{if}\ \phi\ \text{is}\
    \exists x.\ \psi
  \end{cases}
\end{gather*}
Notice, how we never recursively step into the terms. As the name
suggests we only recurse over the logical connectives and quantifiers.
Hence, $g(x)$ is not a subformula of $\lnot f(g(x),c)$ whereas
$f(g(x),c)$ is (since ``$\lnot$'' is a logical connective of
formulas). Importantly, the subformula property preserves the tree
structure of a formula. Hence, formulas with similar sets of
subformulas are related by this property. Therefore, we believe that
recognising this property is important for obtaining a proper
embedding of formulas. In the presented dataset the original formulas
$\phi$ are taken from the TPTP dataset. Unfortunately, finding
negative examples is not as straightforward, since each formula has
infinitely many formulas that are not subformulas. In our dataset, we
only provide the files as described above (positive examples). To
create negative examples during training, we randomly search for
formulas that are not a subformula. Since we want to have
balanced training data we search for as many negative examples as
positive ones.

\paragraph{Modus Ponens:}
One of the most natural logical inference rules is called \emph{modus
  ponens}. The modus ponens (MP) allows the discharging of
implications as shown in the inference
rule~(\ref{prooftree:modus:ponens}). In other words, the consequent
(right-hand side of implication) can be proven to be true if the
antecedent (left-hand side of implication) can be proven.
\begin{equation}\label{prooftree:modus:ponens}
\begin{prooftree}
    \infer0{P}
    \infer0{P \to Q}
    \infer2{Q}
\end{prooftree}
\end{equation}
Using this basic inference rule we associate two formulas $\phi$ and
$\psi$ with each other if $\phi$ can be derived from $\psi$ in few
inference steps with modus ponens and conjunction elimination without unification and matching. 
It turns out that despite its
simplicity, modus ponens makes for a sound and complete proof calculus
for the (undecidable) fragment of first-order logic known as Horn
Formulas~\cite{borger2001classical}.
\begin{example}{}
  We can associate the two formulas
  $\phi := \forall x.\ ((P(x) \to Q(x)) \land P(x))$ and
  $\psi := \forall x.\ Q(x)$ with each other, since $\psi$ can be
  proven from $\phi$ using the modus ponens inference rule (and some
  others).
\end{example}
Providing data for this property required more creativity. We had
two approaches: Option one involves generating data directly from the
TPTP dataset, while the other option comprised synthesising data
ourselves with random strings. In the data set, we provide both
alternatives are used. First, we search for all formulas in the TPTP
set that contained an implication and added the antecedent using a
conjunction. We paired this formula with the formula containing only
the consequent. We tried to introduce heterogeneity to this data by
swapping around conjuncts and even adding other conjuncts in-between.
Secondly, we synthesise data using randomly generated predicate
symbols.

\paragraph{Alpha-Equivalence:}
Two formulas or terms are alpha equivalent if they are equal modulo
variable renaming.
For example, the formulas $\forall x\ y.\ P(x) \land Q(x, y)$ and
$\forall z\ y.\ P(z) \land Q(z, y)$ are alpha equivalent. Alpha
equivalence is an important property for two reasons. First, it
implicitly conveys the notion of variables and their binding. Second,
one often works on alpha equivalence classes of formulas, and hence,
alpha equivalent formulas need to be associated with each other.
\paragraph{Term vs Formula:} We generally want to be able to
distinguish between formulas and terms. This is a fairly simple
property, especially since it can essentially be read off the
BNFs~\ref{eqs:bnf:terms} and~\ref{eqs:bnf:formula}. However, it is
still important to distinguish these two concepts, and a practical
embedding should be able to do so.
\paragraph{Unifiability:} Unifiability plays an important role in many
areas of automated reasoning such as resolution or
narrowing~\cite{Baader:1998:TR:280474}. Unifiability is a property
that only concerns terms. Formally, two terms are unifiable if there
exists a substitution $\sigma$ such that
$s\cdot\sigma \approx t\cdot\sigma$. Informally, a substitution is a
mapping from variables to terms and the application of a substitution
is simply the replacing of variables by the corresponding terms.
Formally one needs to be careful that other variables do not become
bound by substitutions. Example~\ref{ex:unifiability} showcases these
concepts in more detail.
\begin{example}{Substitution and Unifiability:}\label{ex:unifiability}
  The terms $t = f(g(x), y))$ and $ s = f(z,h(0))$ are unifiable,
  since we can apply the substitution:
  $\{z \mapsto g(x),\ y \mapsto h(0)\}$ such that
  $t\cdot\sigma = f(g(x), h(0)) = f(g(x), h(0)) = s\cdot\sigma$.
\end{example}
Syntactic unification, which is the type of unification described
above is quite simple and can be realised with a small set of
inference rules. Note that we only consider the relatively simple syntactic unification problem.
Interestingly, adding additional information such as
associativity or commutativity can make unification an extremely
complex problem~\cite{Baader:1998:TR:280474}. Putting unification into
a higher-order setting makes it even
undecidable~\cite{DBLP:conf/tphol/Huet02}. Both of these problems could be considered in future work.

\subsection{Mizar40 dataset}\label{sec:mizar40_dataset}
Mizar40 dataset~\cite{Kaliszyk2015MizAR4F} is extracted from the mathematical library of the Mizar proof system~\cite{BancerekBGKMNP18}.
The library covers all major domains of mathematics and includes a number of proofs from theorem proving. As such, we believe that it
is representative of the capability of the developed encodings to generalize to theorem proving.
The dataset is structured as follows. Each theorem (goal) is linked to two sets of theorems. One set, the positive examples, are theorems useful in proving the original theorem, and one set, the negative examples, is a set of theorems that were not used in proving the goal.
Note that for each theorem its positive and negative example set are the same size.
The negative examples are selected by a nearest neighbor heuristic\footnote{A more detailed description of the dataset can be found here: \texttt{\url{https://github.com/JUrban/deepmath}}}. Using this data we generate pairs (consisting of a theorem and a premise) and assign them a class based on whether the premise was useful in proving the theorem.

\section{Experiments}\label{sec:eval}
Since the explicit approach does not allow for decoding formulas,
we separately evaluate the two approaches.
We first discuss the evaluation of the explicit approach. We first discuss the performance
of the different encoding models with respect to the properties they were trained with as well as
separate evaluation, where we train a simple model with the resulting encodings.
Then in \prettyref{sec:eval-implicit-approach} we discuss the evaluation of the implicit approach
based on autoencoders. We discuss the decoding accuracy, performance on logical properties discussed previously,
and the theorem proving task of premise selection.

\subsection{Experiments and Evaluation of Explicit Approach}
We will present an evaluation of the explicit encoding models.
First, we consider the properties the models have been
trained with (cf. Section~\ref{sec:prelim}). Here, we have two
different ways of obtaining evaluation and test data. We also want the
encoding networks to generalise to, and preserve properties that it
has not specifically been trained on. Therefore, we encode a set of
formulas and expressions and train an SVM (without kernel
modifications) with different properties on them.

For the first and more straightforward evaluation, we use the data
extracted dataset from the Graph Theory and Set Theory library described in
\prettyref{sec:logical-properties} as training data. One could split this data before training into a training set
and evaluation set so that the network is evaluated on unseen data. In
this approach, however, constants, formulas, etc. occurring in the
evaluation data may have been seen before in different contexts. For
example, considering the Set Theory library, terms and formulas
containing \texttt{union(X,Y)}, \texttt{intersection(X,Y)}, etc. will
occur in training data and evaluation data. Indeed, in applications
such as premise selection, such similarities and connections are
actually desired, which is one of the reasons we use character level
encodings. Nevertheless, we will focus on more difficult
evaluation/test data. We will use data extracted from the Category
Theory library as evaluation data and the Set/Graph Theory data for
training. Hence, training and evaluation sets are significantly
different and share almost no terms, constants, formulas, etc. We train
the models on embedding dimensions 32, 64, and 128 (we only consider
64 for projective models). The input length, i.e. the length of the
formulas was fixed to 256, since this includes almost all training
examples. The CNN models had 8 convolution/pooling layer pairs of
increasing filter sizes (1 to 128), while the LSTM models consisted
of 3 bidirectional LSTM layers each of dimension 256. In the ``Fully
Connected''-models we append two additional dense layers. Similarly,
for the projective models, we append a dense layer with a lower output
dimension.

The evaluation results of the models are shown in
Table~\ref{tab:evaluation}.
\begin{table*}
  \centering \scalebox{0.58}{
    \begin{tabular}{|l|l|l|l|l|l|l|l|l|}
      \hline
      Network                                     & \begin{tabular}[c]{@{}l@{}}embedding \\ dimension\end{tabular} & \begin{tabular}[c]{@{}l@{}}subformula\\ multi-label\\ classification\end{tabular} & \begin{tabular}[c]{@{}l@{}}binary\\ subformula\\ classification\end{tabular} & \begin{tabular}[c]{@{}l@{}}modus \\ ponens\end{tabular} & \begin{tabular}[c]{@{}l@{}}term vs formula\\ classification\end{tabular} & unifiability & \begin{tabular}[c]{@{}l@{}}well-\\ formedness\end{tabular} & \begin{tabular}[c]{@{}l@{}} alpha\\ equivalence\end{tabular} \\ \hline
      CNN                                         & 32  &   0.999&0.625&0.495&0.837&0.858&0.528&0.498 \\ \hline
      CNN                                         & 64  &   0.999&0.635&0.585&0.87&0.73&0.502&0.55 \\ \hline
      CNN                                         & 128 &   0.999&0.59&0.488&0.913&0.815&0.465&0.587 \\ \hline
      CNN with Projection to 32                   & 64  &   1.0&0.662& \cellcolor{blue!25}0.992&0.948&0.81&0.748&0.515 \\ \hline
      CNN with Projection to 64                   & 64  &   1.0&0.653&0.985&0.942&0.718& \cellcolor{blue!25}0.85&0.503 \\ \hline
      CNN with Fully Connected layer              & 32  &   0.999&0.64&0.977&0.968&0.78&0.762&0.5 \\ \hline
      CNN with Fully Connected layer              & 64  &   0.999&0.668&0.975& \cellcolor{blue!25}0.973&0.79&0.77&0.548 \\ \hline
      CNN with Fully Connected layer              & 128 &   0.999&0.635&0.923&0.972&0.828&0.803&0.472 \\ \hline
      CNN with Fully Connected layer Pr to 32     & 64  &   1.0&0.648&0.973&0.922&0.865&0.69&0.487 \\ \hline
      CNN with Fully Connected layer Pr to 64     & 64  &   1.0&0.662&0.968&0.967&0.898&0.762&0.497 \\ \hline
      \rowcolor{Gray}
      LSTM                                        & 32  &   1.0&0.652&0.488&0.975&0.883&0.538&0.508 \\ \hline
      \rowcolor{Gray}
      LSTM                                        & 64  &   0.999&0.652&0.49&0.942&0.86&0.49&0.575 \\ \hline
      \rowcolor{Gray}
      LSTM                                        & 128 &   1.0&0.643&0.473&0.96&0.885&0.51&0.467 \\ \hline
      \rowcolor{Gray}
      LSTM Pr to 32                               & 64  &   1.0& \cellcolor{blue!25}0.69&0.537&0.863&0.87&0.513&0.62 \\ \hline
      \rowcolor{Gray}
      LSTM Pr to 64                               & 64  &   1.0&0.598&0.535&0.845& \cellcolor{blue!25}0.902&0.515&0.575 \\ \hline
      \rowcolor{Gray}
      LSTM with Fully Connected layer             & 32  &   0.999&0.638&0.485&0.855& \cellcolor{blue!25}0.902&0.532&0.692 \\ \hline
      \rowcolor{Gray}
      LSTM with Fully Connected layer             & 64  &   1.0&0.63&0.491&0.882&0.848&0.52& \cellcolor{blue!25}0.833 \\ \hline
      \rowcolor{Gray}
      LSTM with Fully Connected layer             & 128 &   1.0&0.635&0.473&0.968&0.887&0.51&0.715 \\ \hline
      \rowcolor{Gray}
      LSTM with Fully Connected layer Pr to 32    & 64  &   1.0&0.657&0.495&0.96&0.883&0.505&0.672 \\ \hline
      \rowcolor{Gray}
      LSTM with Fully Connected layer Pr to 64    & 64  &   1.0&0.62&0.503&0.712&0.898&0.492&0.662 \\ \hline
    \end{tabular}
  }

  \caption{Accuracies of classifiers working on different
    encoding/embedding models. The models were trained on the
    Graph/Set theory data set and the evaluation was done on the
    unseen Category Theory data set. The LSTM based models are in
    grey. (Pr = Projection)}
  \label{tab:evaluation}
\end{table*}
The multi-label subformula classification is not relevant for this
evaluation since training and testing data are significantly different.
However, the binary subformula classification is useful and proves to
be a difficult property to learn\footnote{The binary subformula classification describes the following problem: Given two formulas, decide if one is a subformula of the other.}. Surprisingly, adding further fully
connected layers seems to have no major effect for this property
regardless of the underlying model. In contrast, the additional dense
layers vastly improve the accuracy of the modus ponens classifier
(from 49\% to 97\% for the simple CNN based model with embedding
dimension 32). It does not make a difference whether these dense
layers are projective or not. Interestingly, every LSTM model even the
ones with dense layers fail when classifying this property. Similar
observations although with a smaller difference can be made with the
term-formula distinction. Classifying whether two terms are unifiable
or not seems to be a task where LSTMs perform better. Generally,
the results for unifiability are similarly good across
models. When determining whether a formula is well-formed, CNN based
models again outperform LSTMs by a long shot. In addition, a big
difference in performance can be seen between CNN models with
additional layers (projective or not) appended. Unsurprisingly alpha
equivalence is a difficult property to learn especially for CNNs. This
is the only property where LSTMs clearly outperform the CNN models. Thus
combining LSTM and CNN layers into a hybrid model might prove beneficial
in future works. In addition, having fully connected layers appears to 
be necessary in order to achieve accuracies significantly above 50\%.

Generally, varying embedding dimensions does not seem to have a great
impact on the performance of a model, regardless of the considered property. As
expected, adding additional fully connected layers has no negative
effect. This leads us to distinguish two types of the properties:
Properties where additional dense layers have a big impact on the
results (modus-ponens, well-formedness, alpha-equivalence), and those
where the effect of additional layers is not significant
(unifiability, term-formula, bin. subformula). It does not seem to
make a big difference whether the appended dense layers are projective
or not. Even the embedding models that embed the formulas to an
\nth{8} of the input dimension perform very well. Another way of
classifying the properties is to group properties where CNNs perform
significantly better (modus-ponens, well-formedness), and conversely
where LSTMs are preferable (alpha equivalence).

\paragraph{Alternative Problems and Properties}
We also want the encodings of formulas to retain information about the
original formulas and properties that the networks have not
specifically been trained on. We want the networks to learn and
preserve unseen structures and relations. We conduct two lightweight
tests for this. First, we train simple models such as SVMs to
recognise certain structural properties such as the existence of certain quantifiers, connectives, etc. (that we did not specifically
train for) in the encodings of formulas. To this end, we train SVMs to
detected logical connectives such as conjunction, disjunction,
implication, etc. These classifications are important since logical
connectives were not specifically used to train the encoding networks but are important nevertheless. Here, the
SVMs correctly predict the presence of conjunctions, etc. with an
accuracy of 85\%. We also train an ordinary linear regression model to
predict the number of occurring universal and existential quantifiers
in the formulas. This regression correctly predicts the number of
quantifiers with an accuracy of 94\% (after rounding to the closest
integer). These results were achieved by using the CNN based model
with fully connected layers. We also evaluated the projective models
with this method. We achieved 70\% and 84\% for classification and
regression respectively using the CNN model with a fully connected and
a projection layer. When using models that were trained using single
layer classifiers as discussed in \prettyref{sec:framework} we get
better results for simple properties such as the presence of
a conjunction.
%
%
\subsection{Experiments and Evaluation of Implicit Approach}\label{sec:eval-implicit-approach}
We also evaluate the encoding models based on the autoencoder setup.
In our experiments, we first learn from unlabelled data.
Hence, we take the entire dataset and discard all labels and simply treat them as formulas. Using this dataset we train encoders and decoders in 100k optimization steps.
First we evaluate how a simple feed-forward network performs when tasked with classifying formulas based on their embeddings.
To this end, we train a feed-forward network to classify input vectors according to properties given in the dataset (logical properties or whether the premise is useful in proving the conjecture). Those input vectors are given by an encoder network whose weights are frozen during this training.
The classifier networks have 6 layers each with size 128 and nonlinear ReLU activation functions.
Since the classification tasks for some properties require two formulas, the input of those classifiers is the concatenated encoding of the input formulas.
We split the classification datasets into training, validation and test sets randomly, in proportions 8-1-1. Every thousand optimization steps we evaluate the validation loss (the loss on the validation set) and report test accuracy from the lowest validation loss point during training.

\paragraph{Hyperparameters} All autoencoding models were trained for 100k steps, using the Adam optimizer \cite{kingma2014adam} with learning rate $1\mathrm{e}-4$, $\beta_1 = 0.9, \beta_2 = 0.999, \epsilon = 1\mathrm{e}-8$. All models work with 128 dimensional sequence token embeddings, and the dimensionality of the final formula encoding was also 128. All models (except for LSTM) are comprised of 6 layers.
In the convolutional network after every convolutional layer, we apply maximum pooling of 2 neighbouring cells.
In the Transformer encoder we use 8 attention heads.
The autoencoders were trained for 100k optimization steps and the classifiers for 30k steps.
The batch size was 32 for difference training, 16 for recursive training, and 32 for classifier networks.
\subsubsection{Decoding accuracy}
\begin{table}[htbp]
\centering
\begin{tabular}{ c c | c | c | c | c }
    & & \multicolumn{2}{| c }{Difference tr.} & \multicolumn{2}{| c }{Recursive tr.} \\
    \hline
    \multicolumn{2}{ c |}{} & Formula & Symbol & Formula & Symbol \\
    \hline
    \multirow{4}{*}{Mizar40 dataset} 
     & Convolutional   & 0.000 & 0.226 & 0.005 & 0.658 \\
     & WaveNet         & 0.000 & 0.197 & 0.006 & 0.657 \\
     & LSTM            & 0.000 & 0.267 & 0.063 & 0.738 \\
     & Transformer     & 0.000 & 0.290 & 0.006 & 0.691 \\
     \hline
     \multirow{4}{*}{Logical properties dataset}
     & Convolutional    & 0.440 & 0.750 & 0.886 & 0.984 \\
     & WaveNet          & 0.420 & 0.729 & 0.865 & 0.981 \\
     & LSTM             & 0.451 & 0.759 & 0.875 & 0.979 \\
     & Transformer      & 0.474 & 0.781 & 0.916 & 0.990 \\
\end{tabular}
\caption{\label{tab:decoding_acc} Decoding accuracy of tested encoders. 
``Formula'' indicates the share of formulas successfully decoded. ``Symbol'' is the average amount of correctly decoded symbols in a formula.}
\end{table}

\begin{figure}[ht]
    \centering
    \includegraphics[width=\textwidth]{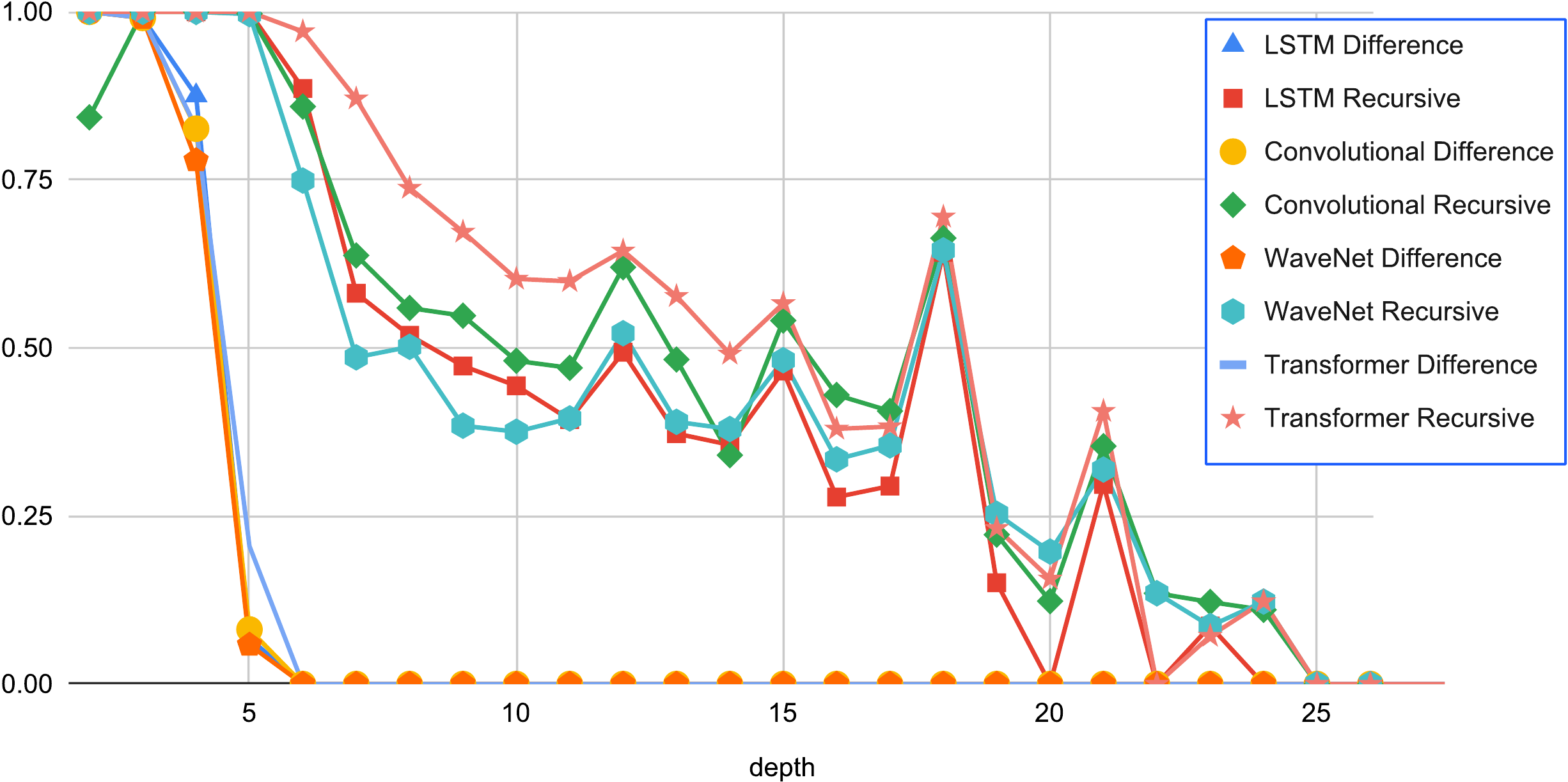}
    \caption{    \label{fig:per_depth_julian}
Decoding accuracy of formulas over formula depth in the logical properties dataset.}
\end{figure}
\begin{figure}[ht]
    \centering
    \includegraphics[width=\textwidth,height=.45\textheight]{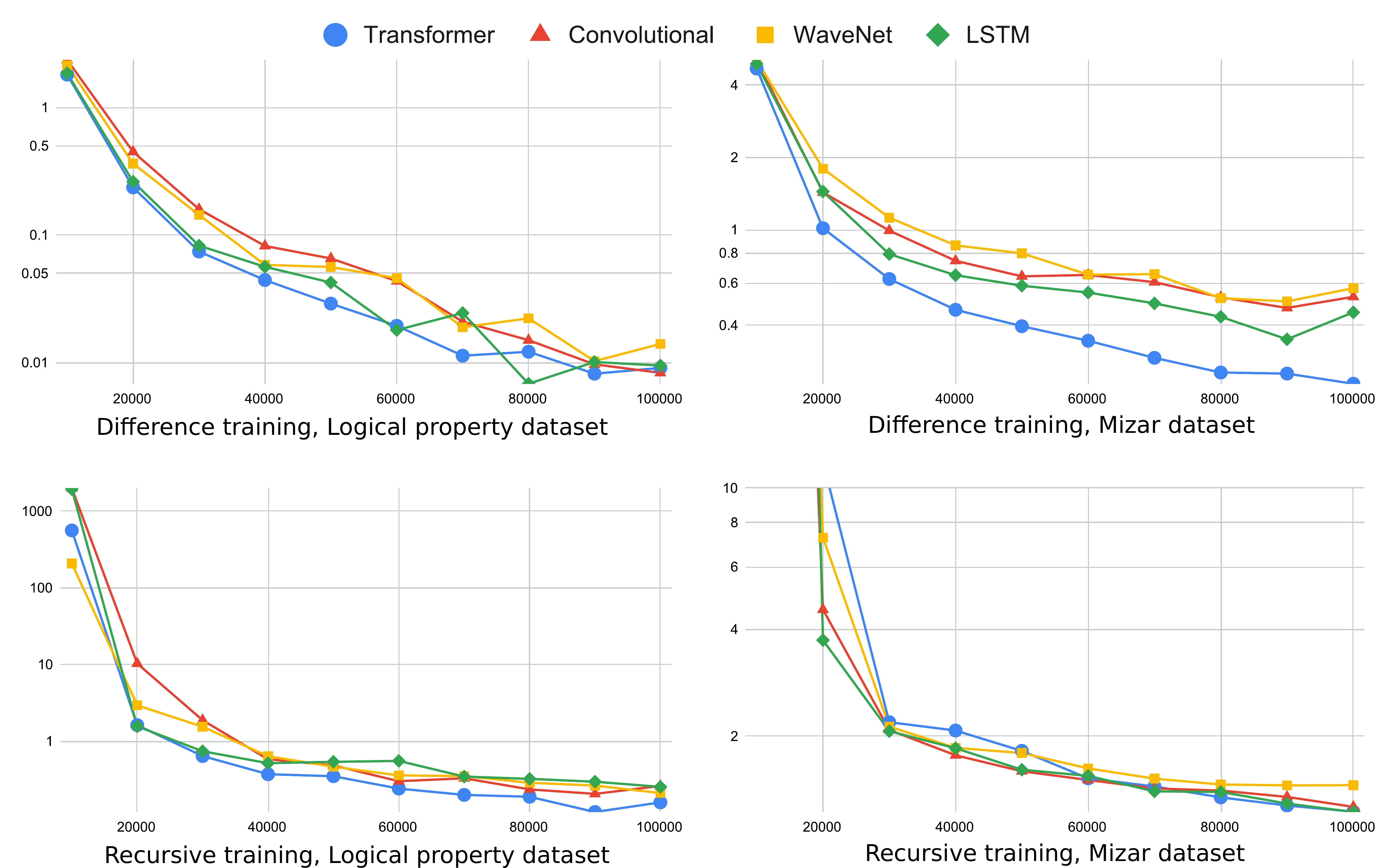}
    \caption{    \label{fig:loss_diff_julian}
Loss during training. The top two graphs present loss during difference training, and the bottom two graphs during recursive training. Note a different vertical scale for the four graphs, this is because the losses for the different training modes and datasets are hard to compare, however all four converge well.}
\end{figure}

After training the autoencoders (\prettyref{fig:loss_diff_julian}) using the unlabelled datasets we test their accuracy. That is, we determine how well the decoder can retrieve the original formulas. This is done recursively. First, the formula is encoded, then its top symbol is determined by the top symbol classifier and encodings of its subformulae are determined using subtree extractors. Then top symbols of those subformulae are found and so on.
The results are presented in \prettyref{tab:decoding_acc}. From the table, it is clear that the recursive training outperforms the difference training regardless of the encoding model or dataset. This result is not unexpected as the design of the recursive training
is more considerate of the subformulas (i.e. subtrees). Hence, a wrong subtree prediction has a larger impact in the loss of
the recursive training than in the difference training. \prettyref{fig:per_depth_julian} shows a plot of the decoding accuracy as
the depth of the formula increases. Unsurprisingly, for very shallow formulas both types of networks perform comparably,
with the difference training accuracy dropping to almost zero as the formulas reach depths 5. On the other hand, the recursive models can almost perfectly recover formulas up to depth 5, which was our goal.

%
%
\subsubsection{Logical properties}\label{sec:eval-logical-properties}
We also test if the encodings preserve logical properties presented in \prettyref{sec:logical-properties}.
In theory, this information still has to be present in some shape or form, but we want to test whether a commonly used feed-forward network can learn to extract them.

\begin{table}[htb]
\centering
\begin{tabular}{ c c | c | c }
     & & Difference tr. & Recursive tr. \\
    \hline
    \multirow{4}{*}{Subformula}
    & Convolutional   & 0.736 & 0.870 \\
    & WaveNet         & 0.787 & 0.877 \\
    & LSTM            & 0.755 & 0.891 \\
    & Transformer     & 0.711 & 0.923 \\
    \hline
    \multirow{4}{*}{Modus Ponens}
    & Convolutional   & 0.920 & 0.893 \\
    & WaveNet         & 0.903 & 0.866 \\
    & LSTM            & 0.941 & 0.916 \\
    & Transformer     & 0.498 & 0.946 \\
    \hline
    \multirow{4}{*}{Term vs Formula}
    & Convolutional   & 1.000 & 0.979 \\
    & WaveNet         & 1.000 & 0.990 \\
    & LSTM            & 1.000 & 0.995 \\
    & Transformer     & 1.000 & 1.000 \\
    \hline
    \multirow{4}{*}{Unifiability}
    & Convolutional   & 0.988 & 0.975 \\
    & WaveNet         & 0.991 & 0.991 \\
    & LSTM            & 0.990 & 0.990 \\
    & Transformer     & 0.989 & 0.990 \\
    \hline
    \multirow{4}{*}{Well-formedness}
    & Convolutional   & 0.969 & 0.988 \\
    & WaveNet         & 1.000 & 0.992 \\
    & LSTM            & 1.000 & 1.000 \\
    & Transformer     & 0.996 & 0.996 \\
    \hline
    \multirow{4}{*}{Alpha equivalence}
    & Convolutional   & 0.998 & 0.998 \\
    & WaveNet         & 0.998 & 1.000 \\
    & LSTM            & 0.483 & 1.000 \\
    & Transformer     & 0.990 & 1.000 \\
\end{tabular}
\caption{\label{tab:eval-logical-properities-implicit}
Logical property classification accuracy on test set.}
\end{table}

The results are shown in \prettyref{tab:eval-logical-properities-implicit}.
Comparing the models we notice a surprising result. Indeed, for some properties,
the difference training performs on-par or even better than recursive training. 
This stands in contrast to the decoding accuracy presented previously where the recursive training outperforms the difference training across the board. This is likely due to the fact that some of the properties can be decided based only on the small top part of the tree, which the difference training does learn successfully (see \prettyref{fig:per_depth_julian}).

\subsubsection{Premise selection}

\begin{table}[htb]
\centering
\begin{tabular}{ c | c | c }
     & Difference tr. & Recursive tr. \\
    \hline
     Convolutional   & 0.681 & 0.696 \\
     WaveNet         & 0.676 & 0.696 \\
     LSTM            & 0.665 & 0.703 \\
     Transformer     & 0.670 & 0.704 \\
\end{tabular}
\caption{\label{tab:mizar_acc} Premise selection accuracy on test set.}
\end{table}

As described before premise selection is an important task in interactive and automated theorem proving.
We test the performance of our encodings for the task of premise selection on the Mizar40 dataset (described in \prettyref{sec:mizar40_dataset}). 
The experiment (as described in \prettyref{sec:eval-implicit-approach}) involves first training the encoder layer to create formula embeddings, then training a feed-forward network to classify formulas by their usefulness in constructing a proof.
The results are shown in \prettyref{tab:mizar_acc}.
Our general decodable embeddings are better than the non-neural machine learning models, albeit perform slightly worse than the best classifiers currently in literature (81\%)~\cite{Crouse2019ImprovingGN} (Which are non-decodable and single-purpose).

\section{Conclusion}\label{sec:concl}

We have developed and compared logical formula encodings (embedding) inspired by the way human mathematicians work.
The formulas are represented in an approximate way, namely as dense continuous vectors. The representations additionally allow for the application
of reasoning steps as well as the reconstruction of the original symbolic expression (i.e. formula) that the vector is supposed to represent.
The explicit approach enforces a number of properties that we would like the embedding to preserve. For example, basic structural
properties (subformula property, etc) can be recovered, natural deduction reasoning steps can be recognised, or even unifiability
between formulas can be checked (although with less precision) in the embedding. In the second approach, we propose to autoencode logical formulas. Here,
we want the encoding of formulas to preserve enough information so that the encoded symbolic expression (formula) can be recovered from the embedding alone.
As such sufficient information for the same logical and structural operations must be present. In addition, this also allows the actual
computation of results of the inference steps or unifiers. We considered two different training setups for the autoencoders. One is called difference training and the other recursive training. In order to train and to evaluate the approaches, we developed several
logical property datasets transformed from subsets of the TPTP problem set.

Apart from an evaluation on the TPTP dataset, we also evaluated the approaches on premise selection problems originating from the
whole Mizar Mathematical Library. As expected, both difference and recursive training are less performant on the Mizar 40
dataset than on the logical properties dataset. We know of two reasons for this. First, the Mizar dataset is much bigger,
both when it comes to the number of constants, types, but also the number of formulas and their average sizes. As such, fitting
all the formulas in vectors of the same size is going to be less precise. Second, the formulas in the Mizar dataset are more
uniformly distributed. As we use models with the same numbers and sizes of layers, memorizing parts of the Mizar dataset is
clearly a more complex task. Despite these problems, the results are promising for both the formula reconstruction task and
the original theorem proving tasks like premise selection.

The code of our embedding, the dataset, and the experiments are available at:\\ \centerline{\texttt{\url{http://cl-informatik.uibk.ac.at/users/cek/logcom2020/}}}

Future work could include considering further logical models and their
variants. We have so far focused on first-order logic, however it is
possible to do the same for simple type theory or even more complex
variants of type theory. This would allow us to do the premise selection analysis presented in this work for the libraries of 
more proof assistants.
Finally, the newly developed capability to decode
an embedding of a first-order formula could also be a useful technique
to consider for conjecturing~\cite{tgckju-cicm16} or proof theory exploration~\cite{hipspec}.
Finally, we imagine that then a reversible encoding of logical formulas could  improve the
proof guidance of first-order logic theorem provers.

\subsection* {Acknowledgements}
This work has been supported
by the ERC starting
grant no.\ 714034 \textit{SMART}.

\bibliographystyle{alpha}
\bibliography{local}

\end{document}